\pdfoutput=1

\documentclass[11pt]{article}

\usepackage[final]{acl}
\usepackage{booktabs}
\usepackage{times}
\usepackage{latexsym}
\usepackage{array}
\usepackage[T1]{fontenc}

\usepackage[utf8]{inputenc}

\usepackage{microtype}

\usepackage{inconsolata}

\usepackage{graphicx}

\usepackage[utf8]{inputenc} 
\usepackage[T1]{fontenc}    
\usepackage{hyperref}       
\usepackage{url}            
\usepackage{booktabs}       
\usepackage{amsfonts}       
\usepackage{nicefrac}       
\usepackage{microtype}      
\usepackage{xcolor}         
\usepackage{color,colortbl}         
\usepackage{times}
\usepackage{adjustbox}
\usepackage{epsfig}
\usepackage{graphicx}
\usepackage{amsmath}
\usepackage{amssymb}
\usepackage{booktabs}
\usepackage{float}
\usepackage{caption}
\usepackage{subcaption}
\usepackage{multirow} 
\usepackage{array}
\title{Med-Agent}

\NewDocumentCommand{\haoyu}{ mO{} }{\textcolor{red}{\textsuperscript{\textit{haoyu}}\textsf{\textbf{\small[#1]}}}}
\NewDocumentCommand{\zihao}{ mO{} }{\textcolor{red}{\textsuperscript{\textit{zihao}}\textsf{\textbf{\small[#1]}}}}
\NewDocumentCommand{\yixin}{ mO{} }{\textcolor{orange}{\textsuperscript{\textit{yixin}}\textsf{\textbf{\small[#1]}}}}

\NewDocumentCommand{\binxu}{ mO{} }{\textcolor{blue}{\textsuperscript{\textit{binxu}}\textsf{\textbf{\small[#1]}}}}

\title{
MMedAgent: Learning to Use Medical Tools with Multi-modal Agent
}

\newcommand{\ie}{\textit{i.e.}}
\newcommand{\eg}{\textit{e.g.}}

\author{
  \textbf{Binxu Li\textsuperscript{1}},
  \textbf{Tiankai Yan\textsuperscript{1}},
  \textbf{Yuanting Pan\textsuperscript{1}},
  \textbf{Jie Luo\textsuperscript{2}},
  \textbf{Ruiyang Ji\textsuperscript{3}},
  \textbf{Jiayuan Ding\textsuperscript{4}},\\
  \textbf{Zhe Xu\textsuperscript{2,5,*}},
  \textbf{Shilong Liu\textsuperscript{6}},
  \textbf{Haoyu Dong\textsuperscript{7}$^{*}$},
  \textbf{Zihao Lin\textsuperscript{3}$^{*}$},
  \textbf{Yixin Wang\textsuperscript{1}\thanks{Corresponding authors }}
\\
\\
  \textsuperscript{1}Stanford University, \quad
  \textsuperscript{2}Harvard Medical School,\quad
  \textsuperscript{3}Virginia Tech,\quad
  \textsuperscript{4}MSU,\quad \\
  \textsuperscript{5}CUHK,\quad 
  \textsuperscript{6}Tsinghua University,\quad
  \textsuperscript{7}Duke University\quad
\\
    {\tt\small \{andy0207,yxinwang\}@stanford.edu, haoyu.dong151@duke.edu, zihaol@vt.edu}\\
}

\begin{document}
\maketitle

\begin{abstract}

Multi-Modal Large Language Models (MLLMs), despite being successful, exhibit limited generality and often fall short when compared to specialized models. 
Recently, LLM-based agents have been developed to address these challenges by selecting appropriate specialized models as tools based on user inputs.
However, such advancements have not been extensively explored within the medical domain. 
To bridge this gap, this paper introduces the first agent explicitly designed for the medical field, named \textbf{M}ulti-modal \textbf{Med}ical \textbf{Agent} (MMedAgent).
We curate an instruction-tuning dataset comprising six medical tools solving seven tasks across
five modalities, enabling the agent to choose the most suitable tools for a given task. 
Comprehensive experiments demonstrate that MMedAgent achieves superior performance across a variety of medical tasks compared to state-of-the-art open-source methods and even the closed-source model, GPT-4o. Furthermore, MMedAgent exhibits efficiency in updating and integrating new medical tools.


\end{abstract}

\section{Introduction}
Multi-modal Large Language Models (MLLMs) have made considerable progress across diverse tasks with inputs from different medical imaging modalities (\eg, Magnetic Resonance Imaging, Computed Tomography, X-ray) in healthcare, including Visual Question Answering (VQA) \cite{moor2023med, zhang2023biomedgpt, llavamed}, image segmentation \cite{medsam}, and Medical Report Generation (MRG) \cite{thawkar2023xraygpt, hamamci2024ct2rep}, etc.
Despite these advancements, MLLMs often exhibit limitations in seamlessly solving multiple tasks across different medical imaging modalities. 
Although recent large medical models \cite{zhang2023biomedclip, tu2024towards, radfm, yang2024advancing, zhao2024biomedparse} have attempted to address this challenge, they remain limited to handling a narrow range of tasks across a restricted set of imaging modalities and cannot be efficiently extended to new tasks or more imaging modalities. Furthermore, these generalists typically do not provide expert-level responses comparable to those of specialized MLLMs customized for specific tasks.


One way to address this issue is to build an AI Agent, an AI system driven by Large Language Models (LLMs) that integrates various domain expert models as tools. Such a system can understand user instructions, make decisions, and select the appropriate tools to execute any specific task, thereby generating expert-level responses for any given request \cite{xie2024large, chen2023llava, wang2024mobile, liu2023llava, tao2023webwise}. 
Despite the significant success of AI agents in the general image domain \cite{tao2023webwise,qin2023mp5,wang2023chatvideo}, there are currently few AI agents developed specifically for the medical domain. Although several works \cite{tang2023medagents, schmidgall2024agentclinic,li2024agent,fan2024ai} in the medical field use the term ``agent'' in their methods, they focus on utilizing LLMs to play various roles and collaborate on complex tasks, in which an ``agent'' refers to a specific role. PathAsst \cite{sun2024pathasst} integrates tool utilization into their framework but specifically designed for pathology tasks.

In this work, we aim to build the first AI agent specifically for the medical domain, termed as \textbf{M}ulti-modal \textbf{Med}ical \textbf{Agent} (MMedAgent). 
We choose LLaVA-Med \cite{llavamed} as the backbone and aim to extend its capability to handle various language and multi-modal tasks, including grounding, segmentation, classification, MRG, and Retrieval-Augmented Generation (RAG). These tasks encompass multiple medical imaging modalities, such as MRI, CT, and X-ray, allowing MMedAgent to support a wide range of data types typically encountered in clinical practice.
The first step to building MMedAgent is to collect the state-of-the-art (SOTA) methods for each task, hereafter referred to as ``tools''. 
During this phase, we identify a lack of an effective tool for the grounding task, prompting us to fine-tune Grounding DINO \cite{groundingdino} specifically for medical applications.
Next, we build an instruction-based dataset that teaches the agent to select the proper tool(s) when encountering a user instruction and aggregate the outputs from tools to reply to users precisely and comprehensively. 
The core of our approach involves an end-to-end training regimen through visual instruction tuning \cite{liu2023llava}.
MMedAgent has demonstrated promising results in various aspects. 
When evaluated on several complex medical tasks, MMedAgent significantly outperforms sevaral open-source SOTA methods, including LLaVA \cite{liu2023improvedllava}, Flamingo-Med \cite{moor2023medflamingo}, Yi-VL-34B \cite{ai2024yi}, Qwen-VL-Chat \cite{Qwen-VL}, LLaVA-Med \cite{llavamed} and RadFM \cite{radfm}, and even surpasses close-source method, GPT-4o \cite{OpenAI2024}, on average. 
It also enhances MMedAgent's backbone, \ie, LLaVA-Med, original capability in the VQA task, as well as exhibits efficient capability in learning new tools. 
Our code and web UI are available at \url{https://github.com/Wangyixinxin/MMedAgent}, and a demonstration of the user interface is provided in Appendix Figure \ref{fig:appendix:server}.

Our contributions can be summarized as:
\begin{itemize}
    \item We propose MMedAgent, the first multi-modal medical AI Agent incorporating a wide spectrum of tools to handle various medical tasks across different modalities seamlessly. 
    \item We build the first open-source instruction tuning dataset for general-purpose multi-modal medical agents. 
    \item Adaptive multi-modal medical tools are incorporated into our Agent. We develop specialized datasets to adapt existing grounding and segmentation tools to the medical domain. 
    \item Extensive experiments demonstrate that MMedAgent surpasses previous SOTA multi-modal medical language models across a range of tasks. 
\end{itemize}
 

\section{Related Work}
\subsection{Medical MLLMs}
LLMs present fertile new ground for research that pushes the frontier of the medical domain. 
Unlike natural domains, the intrinsic complexity of medical data, which includes multiple sources and modalities, has led most LLMs in the medical field to focus on narrowly defined tasks using language and text alone. Singhal et al. \cite{singhal2023large} curate MultiMedQA, a benchmark for medical question-answering datasets, and propose Med-PaLM, which utilizes instruction prompt tuning tailored to medical domains based on PaLM \cite{chowdhery2023palm}. Med-PaLM performs encouragingly on the axes of the human evaluation framework. 

Recent progress on LLMs has been made on multi-modal conversational capability \cite{moor2023med,zhang2023biomedclip,tu2024towards,zhang2023pmc,zhang2023biomedgpt,thawkar2023xraygpt,sun2024pathasst,radfm,llavamed,medsam,yang2024advancing,zhao2024biomedparse,hamamci2024ct2rep}. 
Owing to the diversity inherent in medical data and tasks, LLMs have initially been localized to specific imaging domains such as X-ray \cite{thawkar2023xraygpt}, CT \cite{hamamci2024ct2rep}, and histology \cite{sun2024pathasst}, or tailored for different tasks such as segmentation \cite{medsam, medlsam} and medical report generation \cite{radfm}. In contrast, generalist models expand these capabilities by enabling a single LLM to cover a wider range of imaging modalities and tasks by enlarging the pre-training datasets greatly \cite{zhang2023biomedclip, llavamed, zhao2024biomedparse, liu2023llava, yang2024advancing}. 
Although generalist models are capable of handling a wide range of medical modalities and tasks, they face limitations in scalability when incorporating additional skills and lack specialization in specific tasks. 

\subsection{AI Agent}
A multi-modal AI Agent is a system that achieves users' general-purpose goals by perceiving the environment and making decisions based on the perceptions \cite{xie2024large, wooldridge1995intelligent}. Recent works utilize LLMs as planners to understand multi-modal input from environments and make decisions to call different tools to achieve goals. Based on whether the LLM is open source or not, \citep{xie2024large} classifies multi-modal AI Agents into two types: (i) closed-source LLMs as planners, which utilize prompt technique to enable LLMs to make decisions \cite{chen2023llava, wang2024mobile}; (ii) fine-tuned LLMs as planners, where an LLM is fine-tuned to understand instructions, make decisions, and call tools/APIs \cite{liu2023llava, tao2023webwise, zhang2024toolbehonest}. MMedAgent belongs to the second type.

Multi-modal AI Agents have achieved great success in various applications. For example, \citep{tao2023webwise,gur2023real,zhan2023you} apply agents to control the website or user interface. Some works \cite{qin2023mp5,wang2023jarvis} focus on robotics or embodied AI which applies multi-modal LLMs to perceive and interact with real environments. Most works concentrate on multi-modal understanding, or generation, especially image, video, or audio \cite{liu2023llava,wang2023chatvideo,zhang2023loop}. However, these works are limited to the natural domains.
To the best of our knowledge, we are the first to build a more versatile medical AI Agent, which encompasses a broader spectrum of image modalities, including MRI, CT, X-ray, and histology.


\section{MMedAgent}
Multi-modal Medical Agent (MMedAgent), a system based on an MLLM, is designed to seamlessly manage diverse medical tasks by integrating various open-source medical models. MMedAgent comprises two components: (1) an instruction-tuned multi-modal LLM that functions as an action planner and results aggregator, and (2) a collection of medical tools tailored to the agent, each targeting specific tasks in the medical domain.
We first present the fundamental workflow of MMedAgent in Section \ref{sec:workflow}, followed by a description of creating an instruction-tuning dataset for training the multi-modal LLM as an action planner in Section \ref{sec:instruction-tuning-dataset}. The details of medical tasks and corresponding tools incorporated in MMedAgent are described in Section \ref{sec:multi-modal-medical-tools}.

\subsection{Workflow}
\label{sec:workflow}
\begin{figure}[tbp]
    \centering
    \includegraphics[width=\linewidth]{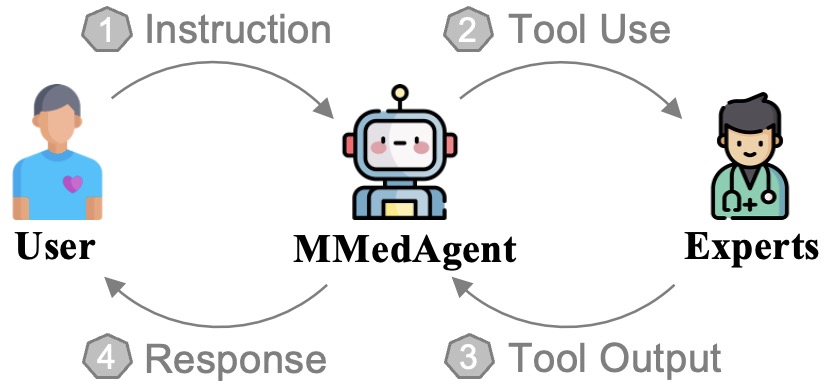}
    \caption{The four-step MMedAgent pipeline.}
    \label{fig:mmedagent}
\end{figure}

Following LLaVA-Plus \cite{liu2023llava}, the objective of MMedAgent is to learn to utilize a wide range of multi-modal medical tools, extending the MLLMs' capabilities to analyze and accomplish various medical tasks. As shown in Figure \ref{fig:mmedagent}, the workflow consists of four parts: (1) users provide an instruction $X_q$ and a medical image $I_q$; (2) MLLM works as an action planner, which understands $X_q$ and $I_q$ and then generates a formatted instruction $X_{\text{tool}}$ to call a specific tool. (3) The tool is executed given $I_q$ and the output $X_{\text{result}}$ of the tool is sent to the MLLM. (4) The MLLM aggregates the output with $X_q$ and $I_q$ and generates the final answer $X_{\text{answer}}$ to users. 
We train the agent end-to-end with an auto-regressive objective on the generated sequence, \textit{i.e.,} $X_{\text{tool}}$ and $X_{\text{answer}}$, to enable the model to use correct tools and answer questions based on the tool's results. 

\begin{figure*}[htbp]
    \centering
    \includegraphics[width=\linewidth, trim={0 0 0 0},clip ]{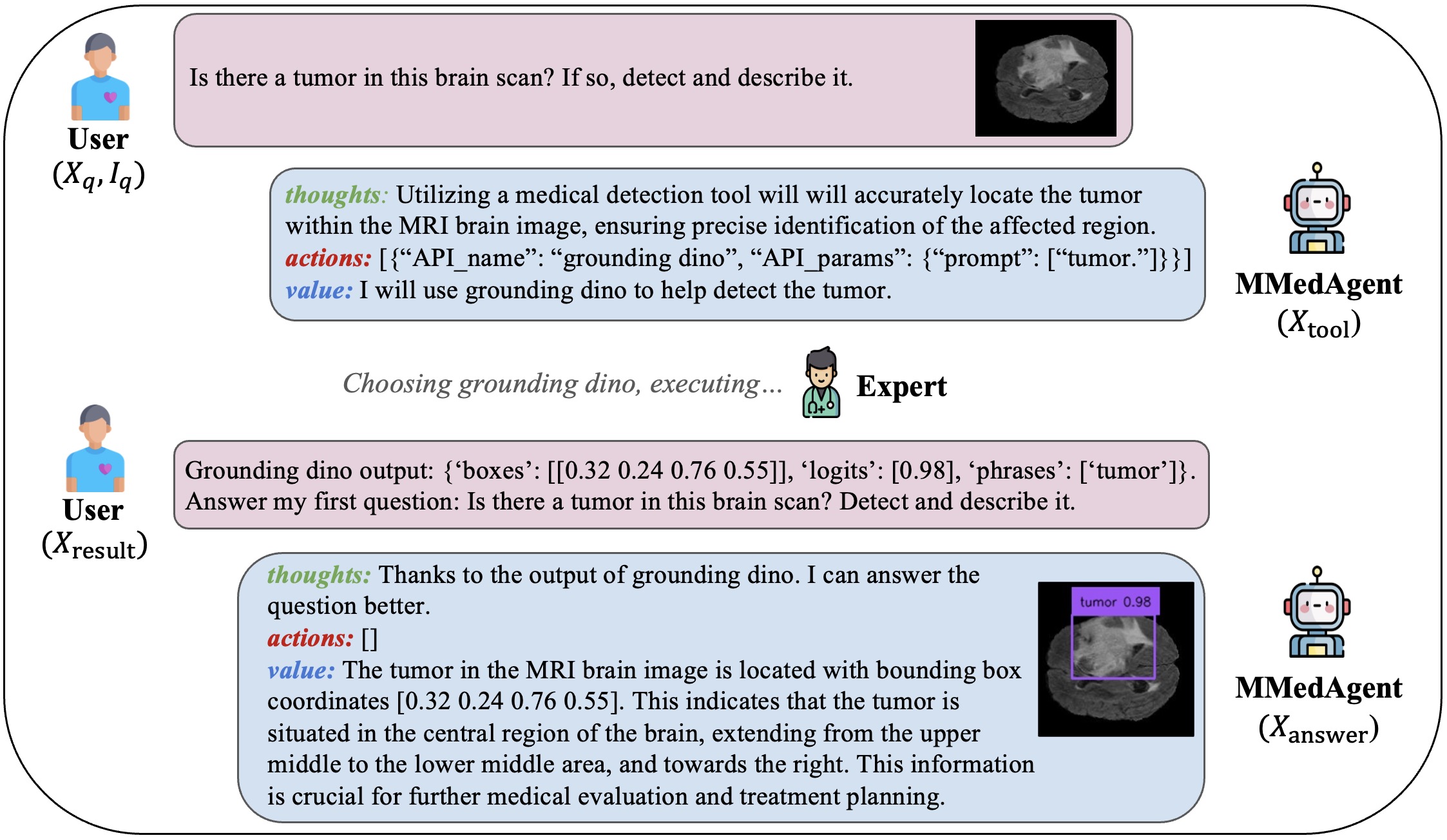}
    \caption{An example of the training data for MMedAgent that learns to use the tool of Grounding DINO for object detection and answer the user's question.}
    \label{fig:instruction-tuning-data}
\end{figure*}
\subsection{Instruction Tuning}
\label{sec:instruction-tuning-dataset}
In order to ensure MMedAgent simultaneously performs as both action planner and results aggregator, we adopt the unified dialogue format proposed by \cite{liu2023llava}, illustrated in Figure \ref{fig:instruction-tuning-data}. Specifically, upon receiving a user's input, MMedAgent generates three components in its outputs: (1) \texttt{Thoughts}, which determine whether MMedAgent can independently solve the user's instructions or if external tools are required, and if so, identifies the appropriate tool; (2) \texttt{Actions}, which enumerate a list of API calls necessary to execute the \texttt{thought}. This comprises two sub-fields: \texttt{API Name} and \texttt{API Params}. If the action list is null, no API call is initiated. (3) \texttt{Value}, which provides a natural language response from the MLLM. During the first round, it indicates the chosen tool(s); during the second round, it represents the final output that addresses the user's initial request. 
As depicted in Appendix Figure \ref{fig:appendix:gpt-instruction-tuning}, we construct the instruction data by querying GPT-4o through one-shot learning, presenting an example that demonstrates the input and output of MMedAgent. We set a fixed \texttt{System} instruction prompt for each tool and select several examples as conversation templates (\texttt{User\_1} and \texttt{Assistant\_1} in Appendix Figure \ref{fig:appendix:gpt-instruction-tuning}). The tool processes the generation of instruction data from the dialogue.

\subsection{Medical Tasks and Tools}
\label{sec:multi-modal-medical-tools}

Our MMedAgent possesses the capability to access a diverse array of tools with the scalability to handle various tasks. As shown in Table \ref{tab:tool_table}, we integrate six tools that encompass seven representative tasks in medical domains, \ie,
(1) grounding, (2) segmentation with bounding-box prompts (B-Seg), (3) segmentation with text prompts (G-Seg), (4) medical imaging classification, (5) Medical Report Generation (MRG), (6) retrieval augmented generation (RAG), and (7) VQA. 
Note that no additional tools are required for the VQA task since we utilize LLaVA-Med, which originally supports the task, as the backbone.
Each tool functions as a specialist, exhibiting exceptional proficiency in executing a specific task across various medical imaging modalities.

\begin{table*}[]
\small
\begin{center}
\begin{tabular}{@{}lllc@{}}
\toprule
Task           & Tool & Data Source                                                                                                                    &Imaging Modality\\ \midrule
VQA            & \begin{tabular}[c]{@{}l@{}}LLaVA-Med\\ \cite{llavamed}\end{tabular}   & \begin{tabular}[c]{@{}l@{}}PMC article\\ \textit{60K-IM}\cite{llavamed}\end{tabular}                                                        & \begin{tabular}[c]{@{}l@{}}MRI, CT, X-ray, \\ Histology, Gross\end{tabular}                                                                  \\\midrule
Classification &  \begin{tabular}[c]{@{}l@{}}BiomedCLIP\\ \cite{biomedclip}\end{tabular}   & \begin{tabular}[c]{@{}l@{}}PMC article\\ \textit{60K-IM}\end{tabular}                                                        & \begin{tabular}[c]{@{}l@{}}MRI, CT, X-ray, \\ Histology, Gross\end{tabular}                                                                  \\\midrule
Grounding      &  \begin{tabular}[c]{@{}l@{}}Grounding DINO \\ \cite{groundingdino}\end{tabular}     & \begin{tabular}[c]{@{}l@{}}WORD, etc.* \\ \end{tabular}   & \begin{tabular}[c]{@{}l@{}}MRI, CT, X-ray, \\ Histology\end{tabular}                                                                 \\\midrule
Segmentation   & \begin{tabular}[c]{@{}l@{}}MedSAM\\ \cite{medsam} \end{tabular}    & WORD, etc.*                                              & \begin{tabular}[c]{@{}l@{}}MRI, CT, X-ray, \\ Histology, Gross\end{tabular}                                                                  \\\midrule
G-Seg.          & \begin{tabular}[c]{@{}l@{}}Grounding DINO\\ + MedSAM \end{tabular}   & \begin{tabular}[c]{@{}l@{}}WORD, etc.* \\ \end{tabular}                                              & \begin{tabular}[c]{@{}l@{}}MRI, CT, X-ray, \\ Histology\end{tabular}                                                                  \\\midrule
MRG            &\begin{tabular}[c]{@{}l@{}}ChatCAD\\ \cite{wang2023chatcad} \end{tabular}   & \begin{tabular}[c]{@{}l@{}}MIMIC-CXR\\ \cite{johnson2019mimiccxrjpg}  \end{tabular}                                                                                                                            & \begin{tabular}[c]{@{}l@{}} X-ray \\\end{tabular}                                                                 \\\midrule
RAG            &\begin{tabular}[c]{@{}l@{}}ChatCAD+\\ \cite{Zhao_2024_chatcad+} \end{tabular}    & \begin{tabular}[c]{@{}l@{}}Merck Manual\\ \cite{Merck_Manual}    \end{tabular}                                                                                                                             & \begin{tabular}[c]{@{}l@{}}-- \\ \end{tabular}                                                                 \\ \bottomrule
\end{tabular}
\end{center}
\vspace{-5pt}
\caption{The tasks, tools, data source, and corresponding medical imaging modalities incorporated in MMedAgent. ``--'' means that the RAG task only focuses on natural language without handling images. 
``WORD, etc.*'' indicates various data sources including WORD \cite{Luo_2022}, FLARE2021 \cite{MedIA-FLARE21}, BRATS \cite{BRATS}, Montgomery County X-ray Set (MC) \cite{chest_cite_1,chest_cite_2}, VinDr-CXR \cite{Nguyen2022}, and Cellseg \cite{NeurIPS-CellSeg}.}
\label{tab:tool_table}
\end{table*}

\subsubsection{Grounding}
\label{sec:grounding-dino}
\noindent\textbf{Grounding}, also known as detection, aims to identify and localize specific objects within an input image by generating the coordinates of bounding boxes containing the objects. To the best of our knowledge, no existing medical models can simultaneously process images from different modalities. Consequently, we propose a generalized grounding tool tailored for the medical domain. Specifically, we choose to fine-tune Grounding DINO \cite{groundingdino}, an open-set object detector, to the medical imaging field.

Our first step is to collect multiple medical image segmentation datasets, including FLARE2021 \cite{MedIA-FLARE21}, WORD \cite{Luo_2022}, BRATS \cite{BRATS}, Montgomery County X-ray Set (MC) \cite{chest_cite_1,chest_cite_2}, VinDr-CXR \cite{Nguyen2022}, and multi-modal cell segmentation dataset (Cellseg) \cite{NeurIPS-CellSeg}. As detailed in Appendix Table \ref{tab:dino-combine_dataset}, these datasets target different modalities, organs, or diseases, each including the original imaging along with their corresponding pixel-level segmentation annotations. These segmentation masks are further transformed into bounding boxes by extracting the minimal outer rectangle around each object. The coordinates of the bounding boxes and the corresponding object labels are then recorded as the grounding labels in each dataset.

Based on the released pre-trained weights, we fine-tuned the Grounding DINO with the dataset described above as well as two common datasets in the natural image field, \ie, COCO \cite{coco} and Flickr30k \cite{flickr30k}, to maintain model's ability in detecting common objects.

\subsubsection{Other Tasks}

\noindent\textbf{B-Seg} involves identifying and delineating the region of interest (ROIs) of an image when a bounding box that covers the ROIs is provided.
It is a type of interactive segmentation, which has become popular since the development of Segment Anything (SAM) \cite{kirillov2023segment}.
We select MedSAM \cite{medsam}, which fine-tunes SAM to the medical field, as our tool.
The prompts are limited to bounding boxes because they provide more precise guidance to SAM \cite{mazurowski2023segment}.
Specifically, in this scenario, we consider the users to provide the position of the bounding box in which MedSAM can be directly applied to obtain the ROI masks. 

\noindent\textbf{G-Seg} refers to combining grounding with SAM. It aims to address a more common scenario when users specify only a particular object to segment in an image. We first activate the fine-tuned grounding tool to localize the referred object and then provide its location, in box format, to MedSAM. Note that this task also presents MMedAgent's capability to address complex tasks with more than one tools, an ability not presented in previous works. 

\noindent\textbf{Classification} aims to identify the most appropriate category for a medical image within a closed set. Specifically, we define a closed set of labels $L$, including organ types, common image modalities, and complex modalities such as ultrasound imaging, hematoxylin, and eosin histopathology. The details of the set $L$ are shown in Appendix \ref{sec:appx-cls}. We adopt BiomedCLIP \cite{biomedclip}, which exhibits superior performance in zero-shot and fine-grained classification.
The image is classified based on the cosine similarity between the image embedding and each text embedding.

\noindent\textbf{MRG} involves creating accurate and authentic medical reports from provided medical information or imaging. MMedAgent incorporates ChatCAD \cite{wang2023chatcad}, an open-source tool designed for generating medical reports for chest X-ray images. 
The model was trained on the MIMIC-CXR dataset \cite{johnson2019mimiccxrjpg} and can provide reports with detailed radiographic analyses, identifying chest-related conditions such as cardiomegaly, edema, consolidation, atelectasis, etc.
\label{sec:4.2}

\noindent\textbf{RAG}\label{sec:retrieval}
refers to enhancing the generated outputs by incorporating the most relevant information acquired from external data sources.
We select ChatCAD+ \cite{Zhao_2024_chatcad+} to implement the medical retrieval process. ChatCAD+ retrieves information from a medical dictionary containing detailed descriptions of 1972 diseases and medical procedures, including their introduction, symptoms, diagnosis, treatment, and causes, sourced from the Merck Manual \cite{Merck_Manual}, a professional medical reference. Given the users' input, the model searches for medical entrees that share the highest cosine similarity with the encoded message and retrieves the relevant knowledge from the medical dictionary. 

\begin{table*}[htbp]
\begin{center}
\label{tab:results}
\setlength{\tabcolsep}{1.4mm}{\begin{tabular}{lcccccccc}
\toprule
     & \multicolumn{3}{c}{Grounding}  & Cls. & MRG  &RAG & Overall & \textcolor{black}{Abs.}\\ 
\cmidrule(lr){2-4} 
& Cell   & Organ  & Disease  \\ \hline
Flamingo-Med \cite{moor2023medflamingo}	&13.11	&15.87	&15.33	&23.56	&16.59 & - & 14.68 & \textcolor{black}{1.16}\\
RadFM \cite{radfm}           & -         & -     & -    & 25.00      & 68.13         & -     & 45.38 & \textcolor{black}{3.59}\\ 
LLaVA-Med \cite{llavamed}          & 51.78     & 65.48     & 68.58   & 53.46    &  70.10    & 30.44     & 60.68 & \textcolor{black}{4.80} \\
Yi-VL-34B \cite{ai2024yi}	&63.23	&79.40	&68.32	&76.02	&72.95	&14.67	&64.08 & \textcolor{black}{5.07} \\
LLaVA-Med (Tool in Test) & 45.32      & 52.77        & 67.91   & 57.53   & 74.34   & 67.55       & 65.31 & \textcolor{black}{5.17} \\ 
Qwen-VL-Chat \cite{Qwen-VL}	&61.34	&65.90	&62.38	&88.40	&73.41	&78.80	&76.21 & \textcolor{black}{6.03}\\
LLaVA-34B \cite{liu2023improvedllava}	&76.75	&84.85	&80.75	&\textbf{96.04}	&80.27	&\textbf{91.64}	&86.52 & \textcolor{black}{6.84} \\

MMedAgent (ours)          &  \textbf{97.50}         & \textbf{102.29}         & \textbf{125.89}  & 81.11  & \textbf{121.49}   & 85.55     & \textbf{109.48} & \textbf{\textcolor{black}{8.66}} \\ \bottomrule
\end{tabular}}
\vspace{-5pt}
\caption{Performance comparison between MMedAgent and other baselines. Cls. stands for classification. We report the relative scores for all tasks and the absolute (abs.) scores for overall performance in the last column. 
``-'' indicates the tasks that the corresponding model is not applicable to. LLaVA-Med refers to the \textit{60K-IM} version with only the initial query $X_q$ and image $I_q$ as input, while LLaVA-Med (Tool in Test) takes $X_q$, $I_q$ and also the internal output from tools $X_{\text{result}}$ as input.}
\label{tab:medagent-bench}
\end{center}
\end{table*}

\section{Experimental Settings}
\label{sec:experimental_setting}
MMedAgent is initialized with LLaVA-Med \textit{60K-IM}, instruction-tuned using LoRA \cite{hu2021lora} for 15 epochs, and conducted over approximately 72 hours on two 80G NVIDIA A100 GPUs. The rank of LoRA is set to 128, and the training batch size is set to 48. We employ AdamW \cite{adamw} as the optimizer alongside a cosine learning rate schedule peaking at 2e-4. 
We generate 48K instruction-tuning data, consisting of 15K augmented VQA instruction following the method from LLaVA-Plus \cite{liu2023llava} derived from 60K inline mentions \cite{llavamed}, 10K data points for detection, 3K for RAG, 5K each for segmentation, classification, MRG, and G-Seg. Data sources are shown in Table \ref{tab:tool_table}.

\section{Experimentals}

We conduct experiments on MMedAgent to answer three research questions: (1) What is the performance of MMedAgent in addressing diverse medical tasks across various modalities (Section \ref{sec:capable-to-solve-complex-medical-tasks})? (2) Does the instruction-tuned MMedAgent exhibit superior performance in open-ended biomedical dialogue (Section \ref{sec:enhancing-existing-capabilities})? (3) What is the efficiency of MMedAgent in invoking tools or incorporating new tools (Section \ref{sec:scalability})?

\subsection{Various Medical Tasks}
\label{sec:capable-to-solve-complex-medical-tasks}
\subsubsection{Evaluation Criterion}
\label{sec:evaluation_criteria}
To evaluate the performance of MMedAgent on various complex medical tasks, we create an evaluation dataset consisting of 70 diverse questions. For this dataset, we initially select 10 concepts randomly from the Merck Manual for RAG and 60 unseen images of different tasks from respective data sources. These include 10 images each for organ grounding,  disease grounding, and cell grounding, along with 20 X-ray images for MRG and 10 images across various modalities for classification. Notably, the VQA task evaluation is shown in Section \ref{sec:enhancing-existing-capabilities}. Due to the inability to describe the segmentation task linguistically, we provide the qualitative results shown in Section \ref{sec:case-study}. Then we utilize the same prompt as outlined in Section \ref{sec:instruction-tuning-dataset} to generate the instruction-tuning data for evaluation. Subsequently, we separately feed the data into GPT-4o, MMedAgent and other benchmarks to obtain the outputs. GPT-4o is a newly released multimodal model with strong visual understanding capabilities. According to the testing from OpenAI, it surpasses GPT-4 Turbo and has a faster inference speed. Thus, the output from GPT-4o can be viewed as a strong benchmark. All the outputs will be assessed by GPT-4 and rated on a scale from 1 to 10 based on their helpfulness, relevance, accuracy, and level of details. We provide GPT-4 with figure captions and include inline mentions from \textit{60K-IM} for the VQA task. The detailed prompts are illustrated in Figure \ref{fig:appendix:gpt-eval_prompt}. For the MRG task, the reports are taken as captions of the input figures. For detection and other tasks without a caption in the original data, we generate the captions by combining the images with the labels, \textit{e.g.,} ``A CT scan showing the kidney organ.''. 
Based on the output from GPT-4o, we propose an absolute score, \textit{i.e.,} the score output by GPT-4, and a relative score, defined as $S_* / S_{GPT-4o} (\%)$, which indicates the performance change caused by other MLLMs. Here, $S_*$ refers to the score of outputs generated by $*$, with $* \in$ \{MMedAgent, LLaVA-Med,...\}. Specifically, we compare with two MLLMs in the medical field, i.e., Med-Flamingo \cite{moor2023medflamingo} and RadFM \cite{radfm}, as well as three generic MLLM, LLaVA-34B \cite{liu2023improvedllava}, Yi-VL-34B \cite{ai2024yi} and Qwen-VL-Chat \cite{Qwen-VL}. A higher score indicates a superior output quality.
During the evaluation, MMedAgent dynamically selects, activates, and executes tools in real-time, then aggregates the obtained results from these tools to answer questions. 



\begin{figure*}[htp]
    \centering
    \includegraphics[width=\textwidth]{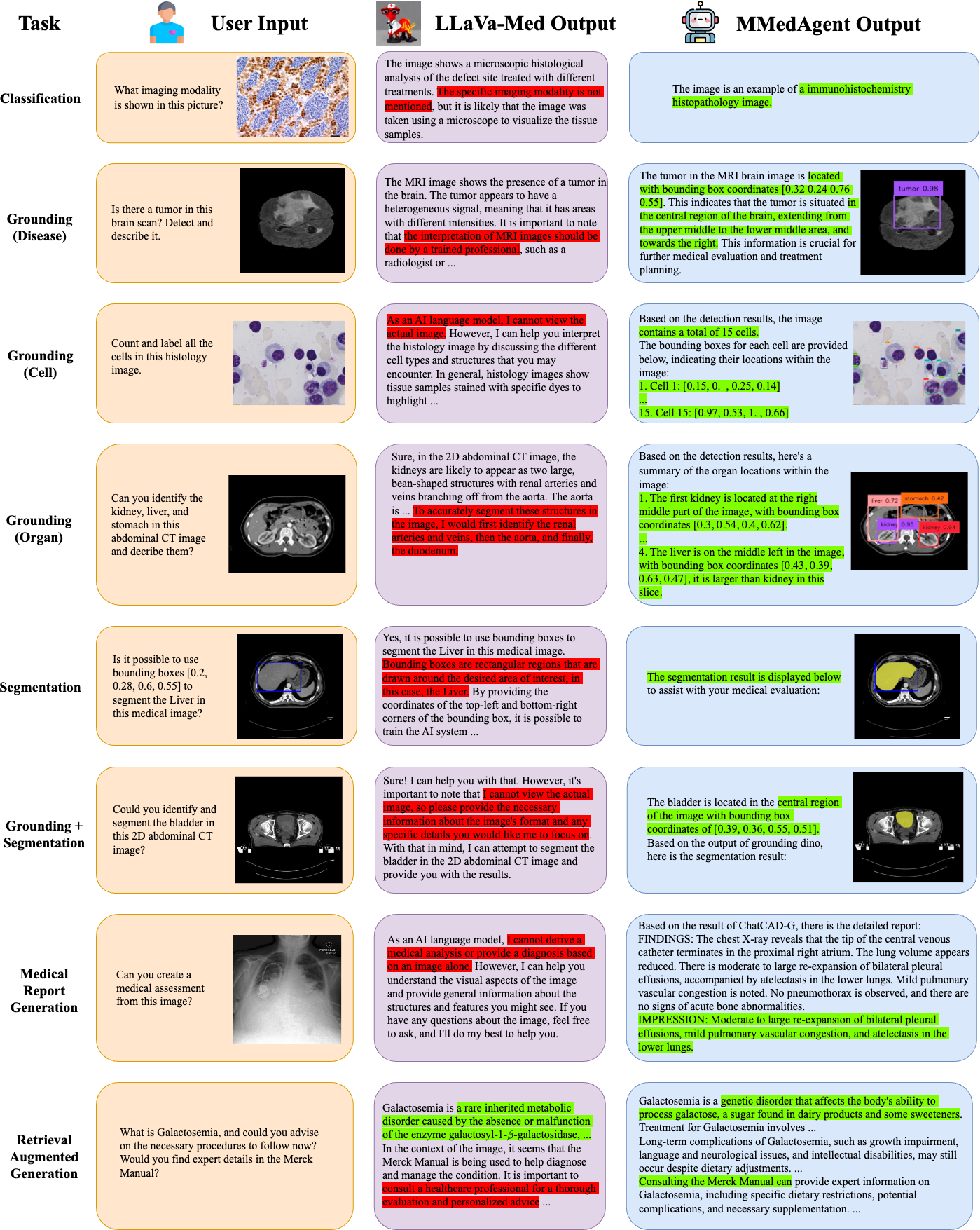}
    \caption{Qualitative comparison between LLaVA-Med and MMedAgent across different tasks. The undesired and desired responses are highlighted in Red and Green respectively.}
    \label{fig:qualitative}
\end{figure*}

\begin{table*}[t!]
    \centering
     \setlength{\tabcolsep}{1.2mm}{
    \begin{tabular}{llcccccccc}
\toprule[1pt]
& & \multicolumn{2}{c}{Question Types} & \multicolumn{5}{c}{Imaging Modalities} & \multirow{2}{*}{Overall} \\ \cmidrule(lr){3-4} \cmidrule(lr){5-9}
& & Conversation & Description & X-ray & MRI & Histology & Gross & CT & \\ 
\multicolumn{2}{l}{(Question Count)} & (143) & (50) & (37) & (38) & (44) & (34) & (40) & (193) \\ \hline
\multicolumn{2}{l}{LLaVA-Med} & 53.30    &   38.90  & 56.58 & \textbf{40.84} & 54.71     & 48.47 & 50.68 & 50.94 \\ 
\multicolumn{2}{l}{MMedAgent} & \textbf{54.49}    &  \textbf{39.75}    & \textbf{58.37} & 35.09 & \textbf{56.88}     & \textbf{51.88} & \textbf{52.79} & \textbf{51.42}\\
\bottomrule[1pt]
\end{tabular}}
\caption{Comparison of open-ended medical dialogue between MMedAgent and LLaVA-Med.}
\label{tab:open-ended-dialogue}
\end{table*}
\begin{table*}[t!]
    \centering
    \setlength{\tabcolsep}{1.7mm}{
    \begin{tabular}{llccccccc}
    \toprule[1pt]
    & & \multicolumn{2}{c}{RAD-VQA} & \multicolumn{2}{c}{SLAKE} & \multicolumn{2}{c}{PATH-VQA} & \multicolumn{1}{c}{PMC-VQA}\\ 
    \cmidrule(lr){3-4} \cmidrule(lr){5-6} \cmidrule(lr){7-8} \cmidrule(lr){9-9}
    & & Open & Close & Open & Close & Open & Close & Close \\ 
    \multicolumn{2}{l}{(Question Count)} & (150) & (150) & (150) & (150) & (150) & (150) & (300) \\ \hline
    \multicolumn{2}{l}{LLaVA-Med} & 28.23 & 61.40 & 39.17 & 52.16 & 12.30 & 54.05 & 27.48  \\ 
    \multicolumn{2}{l}{MMedAgent} & \textbf{58.31} & \textbf{86.72} &  \textbf{79.39} & \textbf{86.34} & \textbf{39.16} & \textbf{90.38}  & \textbf{39.50} \\
    \bottomrule[1pt]
    \end{tabular}}
    \caption{Comparison of VQA tasks between MMedAgent and LLaVA-Med across different VQA benchmarks.}
    \label{tab:open-close-vqa}
\end{table*}

\subsubsection{Experimental Results}
\label{sec:experimental-results}
As illustrated in Table \ref{tab:medagent-bench}, MMedAgent significantly outperforms all other baselines on various tasks. Note that RadFM cannot handle grounding and RAG tasks, and Flamingo-Med is not applicable for RAG because it cannot process text-only input. 
It is observed that the overall relative score of MMedAgent (109.48) outperforms all other state-of-the-art MLLMs by a large margin, being 1.8 times higher than that of LLaVA-Med (60.68), which is the backbone of MMedAgent. We also propose LLaVA-Med (Tool in Test), 
an enhanced version of LLaVA-Med that incorporates the internal output of tools and MMedAgent maintains its superior performance in this case.

Furthermore, the scores for organ grounding, disease grounding, and MRG exceed 100\%, indicating that MMedAgent surpasses GPT-4o in these tasks. These results underscore the superior efficiency of MMedAgent in diverse medical tasks across various modalities.


\subsubsection{Case Study}
\label{sec:case-study}
A detailed visual comparison between LLaVA-Med and MMedAgent is illustrated in Figure \ref{fig:qualitative}.  Given the user queries on tasks involving analyzing the images, such as classification, grounding, and segmentation tasks, LLaVA-Med only generates simple conversational responses without solving the given requests (highlighted in Red) and it is unable to generate visualized results. In contrast, MMedAgent effectively addresses these questions by activating the appropriate tools, integrating their outputs, generating accurate responses (highlighted in Green), and visualizing the results. This is guaranteed by the precise selection of tools by MMedAgent and the superiority of the tools themselves. When encountering language generation-based tasks, \ie, MRG and RAG, LLaVA-Med fails to provide an in-depth analysis of the images. However, MMedAgent provides more straightforward and accurate responses by utilizing the tools designed specifically for these tasks. 
\subsection{Medical VQA}
\label{sec:enhancing-existing-capabilities}
When implementing VQA tasks, MMedAgent can rely on its backbone, i.e., LLaVA-Med (Sections \ref{sec:open-ended-medical-dialogue}) and could be further enhanced by leveraging other VQA specialists (\ref{sec:professional-vqa}). 
\subsubsection{Open-ended Medical Dialogue}
\label{sec:open-ended-medical-dialogue}
We follow the setting of open-ended medical dialogue in LLaVA-Med \cite{llavamed} and use the same test data as LLaVA-Med, consisting of 193 novel questions and 50 unseen images from PMC-15M \cite{biomedclip}. This dataset contains 5 modalities and can be divided into two main classes: conversation questions and detailed description questions. We utilize the same relative score in Section \ref{sec:evaluation_criteria} as the evaluation criterion. Since this is a pure language task, we select the output from GPT-4 rather than GPT-4o as the reference score. 

As shown in Table \ref{tab:open-ended-dialogue}, performance is evaluated from the perspective of question types (conversation and description) and image modalities (X-ray, MRI, Histology, Gross and CT). After instruction-tuning on the tool learning dataset, MMedAgent performs better on both types of questions. Moreover, MMedAgent outperforms LLaVA-Med in all domains but MRI, demonstrating the efficiency of MMedAgent in open-ended medical dialogue.

\subsubsection{VQA Benchmark}
\label{sec:professional-vqa}
MMedAgent is also evaluated on four VQA benchmarks, including VQA-RAD \cite{lau2018dataset}, SLAKE \cite{liu2021slake}, PMC-VQA \cite{zhang2023pmcvqa} and PATH-VQA \cite{he2020pathvqa}. 
We construct the instruction-tuning data for each dataset and fine-tune the model with these newly added VQA tools. For each dataset, We randomly select 4K image-text pairs as the training set and 300 pairs for evaluation. Specifically, VQA-RAD, PATH-VQA, and SLAKE each contain 150 samples from the open set and 150 from the closed set, while PMC-VQA comprises 300 closed multi-choice questions. 
Figure \ref{tab:open-close-vqa} indicates that MMedAgent significantly outperforms LLaVA-Med across all VQA benchmarks.
\begin{figure}[t]
    \centering
    \includegraphics[width=0.45\textwidth]{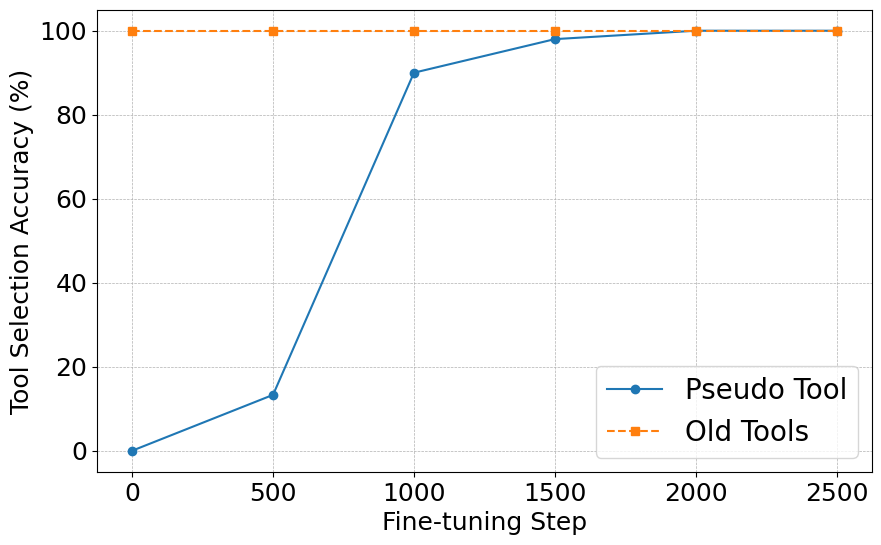}
    \caption{The scalability of MMedAgent.} 
    \label{acc_v_step}
\end{figure}
\subsection{Tool Utilization}
\label{sec:scalability}
The superior performance of MMedAgent on the various tasks described above depends on accurately understanding users' inputs and activating the correct tools. After training MMedAgent for 15 epochs, the tool selection accuracy reached 100\%, demonstrating MMedAgent's ability to select the appropriate tools without errors.

One significance of MMedAgent is its ability to adapt to new tools. 
Here, we consider two scenarios. 
Firstly, when a superior tool for tasks that MMedAgent is already equipped to handle becomes available, the API name of the outdated tool can be seamlessly replaced with that of the new tool, eliminating the need for additional retraining. Secondly, to extend MMedAgent to a new task, it is sufficient to generate a small set of instruction-tuning data for this specific task and fine-tune the agent accordingly, rather than retraining it from the beginning.
To verify this capability, we simulate a new tool called ``Pseudo Tool'', generate an additional 5K instruction-tuning data (following Section \ref{sec:experimental_setting}), and create 30 unseen diverse questions for evaluation following Section \ref{sec:evaluation_criteria}. We utilize the same training settings to fine-tune MMedAgent with a smaller learning rate of 1e-6 and a batch size of 10 on one 80G A100 GPU. As shown in Figure \ref{acc_v_step}, the accuracy of selecting a new tool increase to 100\% within 2K steps without damaging the performance on  selecting old tools.


\section{Conclusion}
We propose MMedAgent, the first multi-modal medical AI agent that is capable of seamlessly utilizing various medical tools to handle a broad spectrum of medical tasks across different imaging modalities. We create an instruction-tuning dataset that MMedAgent utilize to learn to invoke various medical tools and aggregate results from tools. Comprehensive experiments demonstrate that MMedAgent significantly outperforms open-source baselines and even surpasses GPT-4o across many medical tasks. Furthermore, MMedAgent efficiently integrates with new tools while remaining the capability to activate previously learned tools.

\section{Limitation}
Our work is currently limited to seven tasks across five modalities. Due to the need for extensive domain knowledge, 
more specialized tools should be included and MMedAgent's scalability allows for the inclusion of more powerful tools in the future.
Additionally, more generalist LLMs in the medical domain could potentially serve as stronger backbone to enhance MMedAgent.

{\small
\bibliography{MMedAgent_camera_ready}

\begin{thebibliography}{61}
\providecommand{\natexlab}[1]{#1}

\bibitem[{AI et~al.(2024)AI, :, Young, Chen, Li, Huang, Zhang, Zhang, Li, Zhu, Chen, Chang, Yu, Liu, Liu, Yue, Yang, Yang, Yu, Xie, Huang, Hu, Ren, Niu, Nie, Xu, Liu, Wang, Cai, Gu, Liu, and Dai}]{ai2024yi}
01. AI, :, Alex Young, Bei Chen, Chao Li, Chengen Huang, Ge~Zhang, Guanwei Zhang, Heng Li, Jiangcheng Zhu, Jianqun Chen, Jing Chang, Kaidong Yu, Peng Liu, Qiang Liu, Shawn Yue, Senbin Yang, Shiming Yang, Tao Yu, Wen Xie, Wenhao Huang, Xiaohui Hu, Xiaoyi Ren, Xinyao Niu, Pengcheng Nie, Yuchi Xu, Yudong Liu, Yue Wang, Yuxuan Cai, Zhenyu Gu, Zhiyuan Liu, and Zonghong Dai. 2024.
\newblock \href {https://arxiv.org/abs/2403.04652} {Yi: Open foundation models by 01.ai}.
\newblock \emph{Preprint}, arXiv:2403.04652.

\bibitem[{Bai et~al.(2023)Bai, Bai, Yang, Wang, Tan, Wang, Lin, Zhou, and Zhou}]{Qwen-VL}
Jinze Bai, Shuai Bai, Shusheng Yang, Shijie Wang, Sinan Tan, Peng Wang, Junyang Lin, Chang Zhou, and Jingren Zhou. 2023.
\newblock Qwen-vl: A versatile vision-language model for understanding, localization, text reading, and beyond.
\newblock \emph{arXiv preprint arXiv:2308.12966}.

\bibitem[{Candemir et~al.(2014)Candemir, Jaeger, Palaniappan, Musco, Singh, Xue, Karargyris, Antani, Thoma, and McDonald}]{chest_cite_2}
Sema Candemir, Stefan Jaeger, Kannappan Palaniappan, Jonathan~P. Musco, Rahul~K. Singh, Zhiyun Xue, Alexandros Karargyris, Sameer Antani, George Thoma, and Clement~J. McDonald. 2014.
\newblock \href {https://doi.org/10.1109/TMI.2013.2290491} {Lung segmentation in chest radiographs using anatomical atlases with nonrigid registration}.
\newblock \emph{IEEE Transactions on Medical Imaging}, 33(2):577--590.

\bibitem[{Chen et~al.(2023)Chen, Spiridonova, Yang, Gao, and Li}]{chen2023llava}
Wei-Ge Chen, Irina Spiridonova, Jianwei Yang, Jianfeng Gao, and Chunyuan Li. 2023.
\newblock Llava-interactive: An all-in-one demo for image chat, segmentation, generation and editing.
\newblock \emph{arXiv preprint arXiv:2311.00571}.

\bibitem[{Chowdhery et~al.(2023)Chowdhery, Narang, Devlin, Bosma, Mishra, Roberts, Barham, Chung, Sutton, Gehrmann et~al.}]{chowdhery2023palm}
Aakanksha Chowdhery, Sharan Narang, Jacob Devlin, Maarten Bosma, Gaurav Mishra, Adam Roberts, Paul Barham, Hyung~Won Chung, Charles Sutton, Sebastian Gehrmann, et~al. 2023.
\newblock Palm: Scaling language modeling with pathways.
\newblock \emph{Journal of Machine Learning Research}, 24(240):1--113.

\bibitem[{Fan et~al.(2024)Fan, Tang, Chen, Wang, Wei, Xi, Huang, and Zhou}]{fan2024ai}
Zhihao Fan, Jialong Tang, Wei Chen, Siyuan Wang, Zhongyu Wei, Jun Xi, Fei Huang, and Jingren Zhou. 2024.
\newblock Ai hospital: Interactive evaluation and collaboration of llms as intern doctors for clinical diagnosis.
\newblock \emph{arXiv preprint arXiv:2402.09742}.

\bibitem[{Gur et~al.(2023)Gur, Furuta, Huang, Safdari, Matsuo, Eck, and Faust}]{gur2023real}
Izzeddin Gur, Hiroki Furuta, Austin Huang, Mustafa Safdari, Yutaka Matsuo, Douglas Eck, and Aleksandra Faust. 2023.
\newblock A real-world webagent with planning, long context understanding, and program synthesis.
\newblock \emph{arXiv preprint arXiv:2307.12856}.

\bibitem[{Hamamci et~al.(2024)Hamamci, Er, and Menze}]{hamamci2024ct2rep}
Ibrahim~Ethem Hamamci, Sezgin Er, and Bjoern Menze. 2024.
\newblock Ct2rep: Automated radiology report generation for 3d medical imaging.
\newblock \emph{arXiv preprint arXiv:2403.06801}.

\bibitem[{He et~al.(2020)He, Zhang, Mou, Xing, and Xie}]{he2020pathvqa}
Xuehai He, Yichen Zhang, Luntian Mou, Eric Xing, and Pengtao Xie. 2020.
\newblock Pathvqa: 30000+ questions for medical visual question answering.
\newblock \emph{arXiv preprint arXiv:2003.10286}.

\bibitem[{Hu et~al.(2021)Hu, Shen, Wallis, Allen-Zhu, Li, Wang, Wang, and Chen}]{hu2021lora}
Edward~J Hu, Yelong Shen, Phillip Wallis, Zeyuan Allen-Zhu, Yuanzhi Li, Shean Wang, Lu~Wang, and Weizhu Chen. 2021.
\newblock Lora: Low-rank adaptation of large language models.
\newblock \emph{arXiv preprint arXiv:2106.09685}.

\bibitem[{Jaeger et~al.(2014)Jaeger, Karargyris, Candemir, Folio, Siegelman, Callaghan, Xue, Palaniappan, Singh, Antani, Thoma, Wang, Lu, and McDonald}]{chest_cite_1}
Stefan Jaeger, Alexandros Karargyris, Sema Candemir, Les Folio, Jenifer Siegelman, Fiona Callaghan, Zhiyun Xue, Kannappan Palaniappan, Rahul~K. Singh, Sameer Antani, George Thoma, Yi-Xiang Wang, Pu-Xuan Lu, and Clement~J. McDonald. 2014.
\newblock \href {https://doi.org/10.1109/TMI.2013.2284099} {Automatic tuberculosis screening using chest radiographs}.
\newblock \emph{IEEE Transactions on Medical Imaging}, 33(2):233--245.

\bibitem[{Johnson et~al.(2019)Johnson, Pollard, Greenbaum, Lungren, ying Deng, Peng, Lu, Mark, Berkowitz, and Horng}]{johnson2019mimiccxrjpg}
Alistair E.~W. Johnson, Tom~J. Pollard, Nathaniel~R. Greenbaum, Matthew~P. Lungren, Chih ying Deng, Yifan Peng, Zhiyong Lu, Roger~G. Mark, Seth~J. Berkowitz, and Steven Horng. 2019.
\newblock \href {https://arxiv.org/abs/1901.07042} {Mimic-cxr-jpg, a large publicly available database of labeled chest radiographs}.
\newblock \emph{Preprint}, arXiv:1901.07042.

\bibitem[{Kirillov et~al.(2023)Kirillov, Mintun, Ravi, Mao, Rolland, Gustafson, Xiao, Whitehead, Berg, Lo et~al.}]{kirillov2023segment}
Alexander Kirillov, Eric Mintun, Nikhila Ravi, Hanzi Mao, Chloe Rolland, Laura Gustafson, Tete Xiao, Spencer Whitehead, Alexander~C Berg, Wan-Yen Lo, et~al. 2023.
\newblock Segment anything.
\newblock In \emph{Proceedings of the IEEE/CVF International Conference on Computer Vision}, pages 4015--4026.

\bibitem[{Lau et~al.(2018)Lau, Gayen, Ben~Abacha, and Demner-Fushman}]{lau2018dataset}
Jason~J Lau, Soumya Gayen, Asma Ben~Abacha, and Dina Demner-Fushman. 2018.
\newblock A dataset of clinically generated visual questions and answers about radiology images.
\newblock \emph{Scientific data}, 5(1):1--10.

\bibitem[{Lei et~al.(2023)Lei, Wei, Zhang, Li, and Zhang}]{medlsam}
Wenhui Lei, Xu~Wei, Xiaofan Zhang, Kang Li, and Shaoting Zhang. 2023.
\newblock \href {https://arxiv.org/abs/2306.14752} {Medlsam: Localize and segment anything model for 3d ct images}.
\newblock \emph{Preprint}, arXiv:2306.14752.

\bibitem[{Li et~al.(2023)Li, Wong, Zhang, Usuyama, Liu, Yang, Naumann, Poon, and Gao}]{llavamed}
Chunyuan Li, Cliff Wong, Sheng Zhang, Naoto Usuyama, Haotian Liu, Jianwei Yang, Tristan Naumann, Hoifung Poon, and Jianfeng Gao. 2023.
\newblock \href {https://arxiv.org/abs/2306.00890} {Llava-med: Training a large language-and-vision assistant for biomedicine in one day}.
\newblock \emph{Preprint}, arXiv:2306.00890.

\bibitem[{Li et~al.(2024)Li, Wang, Zhang, Li, Lai, Kang, Ma, and Liu}]{li2024agent}
Junkai Li, Siyu Wang, Meng Zhang, Weitao Li, Yunghwei Lai, Xinhui Kang, Weizhi Ma, and Yang Liu. 2024.
\newblock Agent hospital: A simulacrum of hospital with evolvable medical agents.
\newblock \emph{arXiv preprint arXiv:2405.02957}.

\bibitem[{Lin et~al.(2014)Lin, Maire, Belongie, Bourdev, Girshick, Hays, Perona, Ramanan, Zitnick, and Doll{\'a}r}]{coco}
Tsung-Yi Lin, Michael Maire, Serge Belongie, Lubomir Bourdev, Ross Girshick, James Hays, Pietro Perona, Deva Ramanan, C.~Lawrence Zitnick, and Piotr Doll{\'a}r. 2014.
\newblock Microsoft coco: Common objects in context.
\newblock 13.

\bibitem[{Liu et~al.(2021)Liu, Zhan, Xu, Ma, Yang, and Wu}]{liu2021slake}
Bo~Liu, Li-Ming Zhan, Li~Xu, Lin Ma, Yan Yang, and Xiao-Ming Wu. 2021.
\newblock Slake: A semantically-labeled knowledge-enhanced dataset for medical visual question answering.
\newblock In \emph{2021 IEEE 18th International Symposium on Biomedical Imaging (ISBI)}, pages 1650--1654. IEEE.

\bibitem[{Liu et~al.(2023{\natexlab{a}})Liu, Li, Li, and Lee}]{liu2023improvedllava}
Haotian Liu, Chunyuan Li, Yuheng Li, and Yong~Jae Lee. 2023{\natexlab{a}}.
\newblock Improved baselines with visual instruction tuning.

\bibitem[{Liu et~al.(2023{\natexlab{b}})Liu, Cheng, Liu, Zhang, Li, Ren, Zou, Yang, Su, Zhu et~al.}]{liu2023llava}
Shilong Liu, Hao Cheng, Haotian Liu, Hao Zhang, Feng Li, Tianhe Ren, Xueyan Zou, Jianwei Yang, Hang Su, Jun Zhu, et~al. 2023{\natexlab{b}}.
\newblock Llava-plus: Learning to use tools for creating multimodal agents.
\newblock \emph{arXiv preprint arXiv:2311.05437}.

\bibitem[{Liu et~al.(2023{\natexlab{c}})Liu, Zeng, Ren, Li, Zhang, Yang, Li, Yang, Su, Zhu, and Zhang}]{groundingdino}
Shilong Liu, Zhaoyang Zeng, Tianhe Ren, Feng Li, Hao Zhang, Jie Yang, Chunyuan Li, Jianwei Yang, Hang Su, Jun Zhu, and Lei Zhang. 2023{\natexlab{c}}.
\newblock \href {https://arxiv.org/abs/2303.05499} {Grounding dino: Marrying dino with grounded pre-training for open-set object detection}.
\newblock \emph{Preprint}, arXiv:2303.05499.

\bibitem[{Loshchilov and Hutter(2019)}]{adamw}
Ilya Loshchilov and Frank Hutter. 2019.
\newblock \href {https://arxiv.org/abs/1711.05101} {Decoupled weight decay regularization}.
\newblock \emph{Preprint}, arXiv:1711.05101.

\bibitem[{Luo et~al.(2022)Luo, Liao, Xiao, Chen, Song, Zhang, Li, Metaxas, Wang, and Zhang}]{Luo_2022}
Xiangde Luo, Wenjun Liao, Jianghong Xiao, Jieneng Chen, Tao Song, Xiaofan Zhang, Kang Li, Dimitris~N. Metaxas, Guotai Wang, and Shaoting Zhang. 2022.
\newblock \href {https://doi.org/10.1016/j.media.2022.102642} {Word: A large scale dataset, benchmark and clinical applicable study for abdominal organ segmentation from ct image}.
\newblock \emph{Medical Image Analysis}, 82:102642.

\bibitem[{Ma et~al.(2024{\natexlab{a}})Ma, He, Li, Han, You, and Wang}]{medsam}
Jun Ma, Yuting He, Feifei Li, Lin Han, Chenyu You, and Bo~Wang. 2024{\natexlab{a}}.
\newblock \href {https://doi.org/10.1038/s41467-024-44824-z} {Segment anything in medical images}.
\newblock \emph{Nature Communications}, 15(1).

\bibitem[{Ma et~al.(2024{\natexlab{b}})Ma, Xie, Ayyadhury, Ge, Gupta, Gupta, Gu, Zhang, Lee, Kim, Lou, Li, Upschulte, Dickscheid, de~Almeida, Wang, Han, Yang, Labagnara, Gligorovski, Scheder, Rahi, Kempster, Pollitt, Espinosa, Mignot, Middeke, Eckardt, Li, Li, Cai, Bai, Greenwald, Valen, Weisbart, Cimini, Cheung, Brück, Bader, and Wang}]{NeurIPS-CellSeg}
Jun Ma, Ronald Xie, Shamini Ayyadhury, Cheng Ge, Anubha Gupta, Ritu Gupta, Song Gu, Yao Zhang, Gihun Lee, Joonkee Kim, Wei Lou, Haofeng Li, Eric Upschulte, Timo Dickscheid, José~Guilherme de~Almeida, Yixin Wang, Lin Han, Xin Yang, Marco Labagnara, Vojislav Gligorovski, Maxime Scheder, Sahand~Jamal Rahi, Carly Kempster, Alice Pollitt, Leon Espinosa, Tâm Mignot, Jan~Moritz Middeke, Jan-Niklas Eckardt, Wangkai Li, Zhaoyang Li, Xiaochen Cai, Bizhe Bai, Noah~F. Greenwald, David~Van Valen, Erin Weisbart, Beth~A. Cimini, Trevor Cheung, Oscar Brück, Gary~D. Bader, and Bo~Wang. 2024{\natexlab{b}}.
\newblock \href {https://doi.org/10.1038/s41592-024-02233-6} {The multi-modality cell segmentation challenge: Towards universal solutions}.
\newblock \emph{Nature Methods}.

\bibitem[{Ma et~al.(2022)Ma, Zhang, Gu, An, Wang, Ge, Wang, Zhang, Wang, Xu, Gou, Thaler, Payer, Štern, Henderson, McSweeney, Green, Jackson, McIntosh, Nguyen, Qayyum, Conze, Huang, Zhou, Fan, Xiong, Dong, Zhu, He, and Yang}]{MedIA-FLARE21}
Jun Ma, Yao Zhang, Song Gu, Xingle An, Zhihe Wang, Cheng Ge, Congcong Wang, Fan Zhang, Yu~Wang, Yinan Xu, Shuiping Gou, Franz Thaler, Christian Payer, Darko Štern, Edward~G.A. Henderson, Dónal~M. McSweeney, Andrew Green, Price Jackson, Lachlan McIntosh, Quoc-Cuong Nguyen, Abdul Qayyum, Pierre-Henri Conze, Ziyan Huang, Ziqi Zhou, Deng-Ping Fan, Huan Xiong, Guoqiang Dong, Qiongjie Zhu, Jian He, and Xiaoping Yang. 2022.
\newblock Fast and low-gpu-memory abdomen ct organ segmentation: The flare challenge.
\newblock \emph{Medical Image Analysis}, 82:102616.

\bibitem[{Mazurowski et~al.(2023)Mazurowski, Dong, Gu, Yang, Konz, and Zhang}]{mazurowski2023segment}
Maciej~A Mazurowski, Haoyu Dong, Hanxue Gu, Jichen Yang, Nicholas Konz, and Yixin Zhang. 2023.
\newblock Segment anything model for medical image analysis: an experimental study.
\newblock \emph{Medical Image Analysis}, 89:102918.

\bibitem[{Menze et~al.(2015)Menze, Jakab, Bauer, Kalpathy-Cramer, Farahani, Kirby, Burren, Porz, Slotboom, Wiest, Lanczi, Gerstner, Weber, Arbel, Avants, Ayache, Buendia, Collins, Cordier, Corso, Criminisi, Das, Delingette, Demiralp, Durst, Dojat, Doyle, Festa, Forbes, Geremia, Glocker, Golland, Guo, Hamamci, Iftekharuddin, Jena, John, Konukoglu, Lashkari, Mariz, Meier, Pereira, Precup, Price, Raviv, Reza, Ryan, Sarikaya, Schwartz, Shin, Shotton, Silva, Sousa, Subbanna, Szekely, Taylor, Thomas, Tustison, Unal, Vasseur, Wintermark, Ye, Zhao, Zhao, Zikic, Prastawa, Reyes, and Van~Leemput}]{BRATS}
Bjoern~H. Menze, Andras Jakab, Stefan Bauer, Jayashree Kalpathy-Cramer, Keyvan Farahani, Justin Kirby, Yuliya Burren, Nicole Porz, Johannes Slotboom, Roland Wiest, Levente Lanczi, Elizabeth Gerstner, Marc-André Weber, Tal Arbel, Brian~B. Avants, Nicholas Ayache, Patricia Buendia, D.~Louis Collins, Nicolas Cordier, Jason~J. Corso, Antonio Criminisi, Tilak Das, Hervé Delingette, Çağatay Demiralp, Christopher~R. Durst, Michel Dojat, Senan Doyle, Joana Festa, Florence Forbes, Ezequiel Geremia, Ben Glocker, Polina Golland, Xiaotao Guo, Andac Hamamci, Khan~M. Iftekharuddin, Raj Jena, Nigel~M. John, Ender Konukoglu, Danial Lashkari, José~António Mariz, Raphael Meier, Sérgio Pereira, Doina Precup, Stephen~J. Price, Tammy~Riklin Raviv, Syed M.~S. Reza, Michael Ryan, Duygu Sarikaya, Lawrence Schwartz, Hoo-Chang Shin, Jamie Shotton, Carlos~A. Silva, Nuno Sousa, Nagesh~K. Subbanna, Gabor Szekely, Thomas~J. Taylor, Owen~M. Thomas, Nicholas~J. Tustison, Gozde Unal, Flor Vasseur, Max Wintermark, Dong~Hye Ye, Liang
  Zhao, Binsheng Zhao, Darko Zikic, Marcel Prastawa, Mauricio Reyes, and Koen Van~Leemput. 2015.
\newblock \href {https://doi.org/10.1109/TMI.2014.2377694} {The multimodal brain tumor image segmentation benchmark (brats)}.
\newblock \emph{IEEE Transactions on Medical Imaging}, 34(10):1993--2024.

\bibitem[{Moor et~al.(2023{\natexlab{a}})Moor, Huang, Wu, Yasunaga, Dalmia, Leskovec, Zakka, Reis, and Rajpurkar}]{moor2023med}
Michael Moor, Qian Huang, Shirley Wu, Michihiro Yasunaga, Yash Dalmia, Jure Leskovec, Cyril Zakka, Eduardo~Pontes Reis, and Pranav Rajpurkar. 2023{\natexlab{a}}.
\newblock Med-flamingo: a multimodal medical few-shot learner.
\newblock In \emph{Machine Learning for Health (ML4H)}, pages 353--367. PMLR.

\bibitem[{Moor et~al.(2023{\natexlab{b}})Moor, Huang, Wu, Yasunaga, Zakka, Dalmia, Reis, Rajpurkar, and Leskovec}]{moor2023medflamingo}
Michael Moor, Qian Huang, Shirley Wu, Michihiro Yasunaga, Cyril Zakka, Yash Dalmia, Eduardo~Pontes Reis, Pranav Rajpurkar, and Jure Leskovec. 2023{\natexlab{b}}.
\newblock \href {https://arxiv.org/abs/2307.15189} {Med-flamingo: A multimodal medical few-shot learner}.
\newblock ArXiv:2307.15189.

\bibitem[{Nguyen et~al.(2022)Nguyen, Lam, Le, Pham, Tran, Nguyen, Le, Pham, Tong, Dinh, Do, Doan, Nguyen, Nguyen, Nguyen, Hoang, Phan, Nguyen, Ho, Ngo, Nguyen, Nguyen, Dao, and Vu}]{Nguyen2022}
Ha~Q. Nguyen, Khanh Lam, Linh~T. Le, Hieu~H. Pham, Dat~Q. Tran, Dung~B. Nguyen, Dung~D. Le, Chi~M. Pham, Hang T.~T. Tong, Diep~H. Dinh, Cuong~D. Do, Luu~T. Doan, Cuong~N. Nguyen, Binh~T. Nguyen, Que~V. Nguyen, Au~D. Hoang, Hien~N. Phan, Anh~T. Nguyen, Phuong~H. Ho, Dat~T. Ngo, Nghia~T. Nguyen, Nhan~T. Nguyen, Minh Dao, and Van Vu. 2022.
\newblock \href {https://doi.org/10.1038/s41597-022-01498-w} {Vindr-cxr: An open dataset of chest x-rays with radiologist's annotations}.
\newblock \emph{Scientific Data}, 9(1):429.

\bibitem[{OpenAI(2024)}]{OpenAI2024}
OpenAI. 2024.
\newblock Hello gpt-4o.
\newblock \url{https://openai.com/index/hello-gpt-4o/}.
\newblock Accessed: 2024-05-26.

\bibitem[{Plummer et~al.(2015)Plummer, Wang, Cervantes, Caicedo, Hockenmaier, and Lazebnik}]{flickr30k}
Bryan~A. Plummer, Liwei Wang, Chris~M. Cervantes, Juan~C. Caicedo, Julia Hockenmaier, and Svetlana Lazebnik. 2015.
\newblock Flickr30k entities: Collecting region-to-phrase correspondences for richer image-to-sentence models.
\newblock pages 2641--2649.

\bibitem[{Porter and Kaplan(2011)}]{Merck_Manual}
Robert~S. Porter and Justin~L. Kaplan. 2011.
\newblock \emph{The merck manual of diagnosis and therapy, 2011}.
\newblock Merck Research Laboratories.

\bibitem[{Qin et~al.(2023)Qin, Zhou, Liu, Yin, Sheng, Zhang, Qiao, and Shao}]{qin2023mp5}
Yiran Qin, Enshen Zhou, Qichang Liu, Zhenfei Yin, Lu~Sheng, Ruimao Zhang, Yu~Qiao, and Jing Shao. 2023.
\newblock Mp5: A multi-modal open-ended embodied system in minecraft via active perception.
\newblock \emph{arXiv preprint arXiv:2312.07472}.

\bibitem[{Schmidgall et~al.(2024)Schmidgall, Ziaei, Harris, Reis, Jopling, and Moor}]{schmidgall2024agentclinic}
Samuel Schmidgall, Rojin Ziaei, Carl Harris, Eduardo Reis, Jeffrey Jopling, and Michael Moor. 2024.
\newblock Agentclinic: a multimodal agent benchmark to evaluate ai in simulated clinical environments.
\newblock \emph{arXiv preprint arXiv:2405.07960}.

\bibitem[{Singhal et~al.(2023)Singhal, Azizi, Tu, Mahdavi, Wei, Chung, Scales, Tanwani, Cole-Lewis, Pfohl et~al.}]{singhal2023large}
Karan Singhal, Shekoofeh Azizi, Tao Tu, S~Sara Mahdavi, Jason Wei, Hyung~Won Chung, Nathan Scales, Ajay Tanwani, Heather Cole-Lewis, Stephen Pfohl, et~al. 2023.
\newblock Large language models encode clinical knowledge.
\newblock \emph{Nature}, 620(7972):172--180.

\bibitem[{Sun et~al.(2024)Sun, Zhu, Zheng, Zhang, Sun, Shui, Zhang, Li, and Yang}]{sun2024pathasst}
Yuxuan Sun, Chenglu Zhu, Sunyi Zheng, Kai Zhang, Lin Sun, Zhongyi Shui, Yunlong Zhang, Honglin Li, and Lin Yang. 2024.
\newblock Pathasst: A generative foundation ai assistant towards artificial general intelligence of pathology.
\newblock In \emph{Proceedings of the AAAI Conference on Artificial Intelligence}, volume~38, pages 5034--5042.

\bibitem[{Tang et~al.(2023)Tang, Zou, Zhang, Zhao, Zhang, Cohan, and Gerstein}]{tang2023medagents}
Xiangru Tang, Anni Zou, Zhuosheng Zhang, Yilun Zhao, Xingyao Zhang, Arman Cohan, and Mark Gerstein. 2023.
\newblock Medagents: Large language models as collaborators for zero-shot medical reasoning.
\newblock \emph{arXiv preprint arXiv:2311.10537}.

\bibitem[{Tao et~al.(2023)Tao, TV, Shlapentokh-Rothman, Hoiem, and Ji}]{tao2023webwise}
Heyi Tao, Sethuraman TV, Michal Shlapentokh-Rothman, Derek Hoiem, and Heng Ji. 2023.
\newblock Webwise: Web interface control and sequential exploration with large language models.
\newblock \emph{arXiv preprint arXiv:2310.16042}.

\bibitem[{Thawkar et~al.(2023)Thawkar, Shaker, Mullappilly, Cholakkal, Anwer, Khan, Laaksonen, and Khan}]{thawkar2023xraygpt}
Omkar Thawkar, Abdelrahman Shaker, Sahal~Shaji Mullappilly, Hisham Cholakkal, Rao~Muhammad Anwer, Salman Khan, Jorma Laaksonen, and Fahad~Shahbaz Khan. 2023.
\newblock Xraygpt: Chest radiographs summarization using medical vision-language models.
\newblock \emph{arXiv preprint arXiv:2306.07971}.

\bibitem[{Tu et~al.(2024)Tu, Azizi, Driess, Schaekermann, Amin, Chang, Carroll, Lau, Tanno, Ktena et~al.}]{tu2024towards}
Tao Tu, Shekoofeh Azizi, Danny Driess, Mike Schaekermann, Mohamed Amin, Pi-Chuan Chang, Andrew Carroll, Charles Lau, Ryutaro Tanno, Ira Ktena, et~al. 2024.
\newblock Towards generalist biomedical ai.
\newblock \emph{NEJM AI}, 1(3):AIoa2300138.

\bibitem[{Wang et~al.(2023{\natexlab{a}})Wang, Chen, Luo, Dai, Yuan, Wu, and Jiang}]{wang2023chatvideo}
Junke Wang, Dongdong Chen, Chong Luo, Xiyang Dai, Lu~Yuan, Zuxuan Wu, and Yu-Gang Jiang. 2023{\natexlab{a}}.
\newblock Chatvideo: A tracklet-centric multimodal and versatile video understanding system.
\newblock \emph{arXiv preprint arXiv:2304.14407}.

\bibitem[{Wang et~al.(2024)Wang, Xu, Ye, Yan, Shen, Zhang, Huang, and Sang}]{wang2024mobile}
Junyang Wang, Haiyang Xu, Jiabo Ye, Ming Yan, Weizhou Shen, Ji~Zhang, Fei Huang, and Jitao Sang. 2024.
\newblock Mobile-agent: Autonomous multi-modal mobile device agent with visual perception.
\newblock \emph{arXiv preprint arXiv:2401.16158}.

\bibitem[{Wang et~al.(2023{\natexlab{b}})Wang, Zhao, Ouyang, Wang, and Shen}]{wang2023chatcad}
Sheng Wang, Zihao Zhao, Xi~Ouyang, Qian Wang, and Dinggang Shen. 2023{\natexlab{b}}.
\newblock \href {https://arxiv.org/abs/2302.07257} {Chatcad: Interactive computer-aided diagnosis on medical image using large language models}.
\newblock \emph{Preprint}, arXiv:2302.07257.

\bibitem[{Wang et~al.(2023{\natexlab{c}})Wang, Cai, Liu, Jin, Hou, Zhang, Lin, He, Zheng, Yang et~al.}]{wang2023jarvis}
Zihao Wang, Shaofei Cai, Anji Liu, Yonggang Jin, Jinbing Hou, Bowei Zhang, Haowei Lin, Zhaofeng He, Zilong Zheng, Yaodong Yang, et~al. 2023{\natexlab{c}}.
\newblock Jarvis-1: Open-world multi-task agents with memory-augmented multimodal language models.
\newblock \emph{arXiv preprint arXiv:2311.05997}.

\bibitem[{Wooldridge and Jennings(1995)}]{wooldridge1995intelligent}
Michael Wooldridge and Nicholas~R Jennings. 1995.
\newblock Intelligent agents: Theory and practice.
\newblock \emph{The knowledge engineering review}, 10(2):115--152.

\bibitem[{Wu et~al.(2023)Wu, Zhang, Zhang, Wang, and Xie}]{radfm}
Chaoyi Wu, Xiaoman Zhang, Ya~Zhang, Yanfeng Wang, and Weidi Xie. 2023.
\newblock \href {https://arxiv.org/abs/2308.02463} {Towards generalist foundation model for radiology by leveraging web-scale 2d\&3d medical data}.
\newblock \emph{Preprint}, arXiv:2308.02463.

\bibitem[{Xie et~al.(2024)Xie, Chen, Zhang, Wan, and Li}]{xie2024large}
Junlin Xie, Zhihong Chen, Ruifei Zhang, Xiang Wan, and Guanbin Li. 2024.
\newblock Large multimodal agents: A survey.
\newblock \emph{arXiv preprint arXiv:2402.15116}.

\bibitem[{Yang et~al.(2024)Yang, Xu, Sellergren, Kohlberger, Zhou, Ktena, Kiraly, Ahmed, Hormozdiari, Jaroensri et~al.}]{yang2024advancing}
Lin Yang, Shawn Xu, Andrew Sellergren, Timo Kohlberger, Yuchen Zhou, Ira Ktena, Atilla Kiraly, Faruk Ahmed, Farhad Hormozdiari, Tiam Jaroensri, et~al. 2024.
\newblock Advancing multimodal medical capabilities of gemini.
\newblock \emph{arXiv preprint arXiv:2405.03162}.

\bibitem[{Zhan and Zhang(2023)}]{zhan2023you}
Zhuosheng Zhan and Aston Zhang. 2023.
\newblock You only look at screens: Multimodal chain-of-action agents.
\newblock \emph{arXiv preprint arXiv:2309.11436}.

\bibitem[{Zhang et~al.(2023{\natexlab{a}})Zhang, Yu, Yan, Liu, Adhikarla, Fu, Chen, Chen, Zhou, Li et~al.}]{zhang2023biomedgpt}
Kai Zhang, Jun Yu, Zhiling Yan, Yixin Liu, Eashan Adhikarla, Sunyang Fu, Xun Chen, Chen Chen, Yuyin Zhou, Xiang Li, et~al. 2023{\natexlab{a}}.
\newblock Biomedgpt: A unified and generalist biomedical generative pre-trained transformer for vision, language, and multimodal tasks.
\newblock \emph{arXiv preprint arXiv:2305.17100}.

\bibitem[{Zhang et~al.(2024{\natexlab{a}})Zhang, Xu, Usuyama, Xu, Bagga, Tinn, Preston, Rao, Wei, Valluri, Wong, Tupini, Wang, Mazzola, Shukla, Liden, Gao, Lungren, Naumann, Wang, and Poon}]{biomedclip}
Sheng Zhang, Yanbo Xu, Naoto Usuyama, Hanwen Xu, Jaspreet Bagga, Robert Tinn, Sam Preston, Rajesh Rao, Mu~Wei, Naveen Valluri, Cliff Wong, Andrea Tupini, Yu~Wang, Matt Mazzola, Swadheen Shukla, Lars Liden, Jianfeng Gao, Matthew~P. Lungren, Tristan Naumann, Sheng Wang, and Hoifung Poon. 2024{\natexlab{a}}.
\newblock \href {https://arxiv.org/abs/2303.00915} {Biomedclip: a multimodal biomedical foundation model pretrained from fifteen million scientific image-text pairs}.
\newblock \emph{Preprint}, arXiv:2303.00915.

\bibitem[{Zhang et~al.(2023{\natexlab{b}})Zhang, Xu, Usuyama, Xu, Bagga, Tinn, Preston, Rao, Wei, Valluri et~al.}]{zhang2023biomedclip}
Sheng Zhang, Yanbo Xu, Naoto Usuyama, Hanwen Xu, Jaspreet Bagga, Robert Tinn, Sam Preston, Rajesh Rao, Mu~Wei, Naveen Valluri, et~al. 2023{\natexlab{b}}.
\newblock Biomedclip: a multimodal biomedical foundation model pretrained from fifteen million scientific image-text pairs.
\newblock \emph{arXiv preprint arXiv:2303.00915}.

\bibitem[{Zhang et~al.(2023{\natexlab{c}})Zhang, Wu, Zhao, Lin, Zhang, Wang, and Xie}]{zhang2023pmc}
Xiaoman Zhang, Chaoyi Wu, Ziheng Zhao, Weixiong Lin, Ya~Zhang, Yanfeng Wang, and Weidi Xie. 2023{\natexlab{c}}.
\newblock Pmc-vqa: Visual instruction tuning for medical visual question answering.
\newblock \emph{arXiv preprint arXiv:2305.10415}.

\bibitem[{Zhang et~al.(2023{\natexlab{d}})Zhang, Wu, Zhao, Lin, Zhang, Wang, and Xie}]{zhang2023pmcvqa}
Xiaoman Zhang, Chaoyi Wu, Ziheng Zhao, Weixiong Lin, Ya~Zhang, Yanfeng Wang, and Weidi Xie. 2023{\natexlab{d}}.
\newblock Pmc-vqa: Visual instruction tuning for medical visual question answering.
\newblock \emph{arXiv preprint arXiv:2305.10415}.

\bibitem[{Zhang et~al.(2023{\natexlab{e}})Zhang, Maezawa, Xia, Yamamoto, and Dixon}]{zhang2023loop}
Yixiao Zhang, Akira Maezawa, Gus Xia, Kazuhiko Yamamoto, and Simon Dixon. 2023{\natexlab{e}}.
\newblock Loop copilot: Conducting ai ensembles for music generation and iterative editing.
\newblock \emph{arXiv preprint arXiv:2310.12404}.

\bibitem[{Zhang et~al.(2024{\natexlab{b}})Zhang, Chen, Wang, Liu, Yang, Shi, Zhu, Lin, Wan, Yang et~al.}]{zhang2024toolbehonest}
Yuxiang Zhang, Jing Chen, Junjie Wang, Yaxin Liu, Cheng Yang, Chufan Shi, Xinyu Zhu, Zihao Lin, Hanwen Wan, Yujiu Yang, et~al. 2024{\natexlab{b}}.
\newblock Toolbehonest: A multi-level hallucination diagnostic benchmark for tool-augmented large language models.
\newblock \emph{arXiv preprint arXiv:2406.20015}.

\bibitem[{Zhao et~al.(2024{\natexlab{a}})Zhao, Gu, Yang, Usuyama, Lee, Naumann, Gao, Crabtree, Piening, Bifulco et~al.}]{zhao2024biomedparse}
Theodore Zhao, Yu~Gu, Jianwei Yang, Naoto Usuyama, Ho~Hin Lee, Tristan Naumann, Jianfeng Gao, Angela Crabtree, Brian Piening, Carlo Bifulco, et~al. 2024{\natexlab{a}}.
\newblock Biomedparse: a biomedical foundation model for image parsing of everything everywhere all at once.
\newblock \emph{arXiv preprint arXiv:2405.12971}.

\bibitem[{Zhao et~al.(2024{\natexlab{b}})Zhao, Wang, Gu, Zhu, Mei, Zhuang, Cui, Wang, and Shen}]{Zhao_2024_chatcad+}
Zihao Zhao, Sheng Wang, Jinchen Gu, Yitao Zhu, Lanzhuju Mei, Zixu Zhuang, Zhiming Cui, Qian Wang, and Dinggang Shen. 2024{\natexlab{b}}.
\newblock \href {https://doi.org/10.1109/tmi.2024.3398350} {Chatcad+: Towards a universal and reliable interactive cad using llms}.
\newblock \emph{IEEE Transactions on Medical Imaging}, page 1–1.

\end{thebibliography}
}

\appendix
\newpage
\section{Details of Tools}
\label{appendix:details-of-tools}

\subsection{Classification}
\label{sec:appx-cls}
We construct a close set of labels $L$ for BiomedCLIP to search for the most suitable category for the given image.

$L$ =\{``adenocarcinoma histopathology'', ``brain MRI'', ``covid line chart'', ``squamous cell carcinoma histopathology'', ``immunohistochemistry histopathology'', ``bone X-ray'', ``chest X-ray'', ``pie chart'',  ``ultrasound imaging``, ``hematoxylin and eosin histopathology'', ``gross''\}. 

\subsection{Retrieval Augmented Generation (RAG)}
RAG distinguishes itself from standard report generation by its access to an external knowledge base, such as Merck Manual. We consider the following three common uses of RAG. The instruction-tuning data are generated based on these functionalities.



\begin{enumerate}
\item \textbf{Chest X-ray image report analysis.}
The chest X-ray image report analysis can function to analyze the report on medical images and provide an analysis including the potential diseases and their related retrieved knowledge and source. 
\item \textbf{General medical report analysis.}
The general medical report analysis can take a summarized report on common diseases and generate an analysis with medical advice such as treatments and precautions, together with a link to the retrieved source from the Merck Manual official website. 

\item \textbf{General medical advice generation.} 
For general medical advice generation, the user can ask general questions about the diseases, and the model will retrieve and provide related information on them. 
\end{enumerate}
For the chest X-ray image report analysis, we generate 1000 chest X-ray reports from the MRG tool described in Section~\ref{sec:4.2} as the report dataset. For the datasets of general medical report analysis and general medical advice generation, we utilize GPT-4o to generate 1000 medical reports and 1000 patient questions respectively about common diseases sampled from the entrees covered in the Merck Manual.

\subsection{Medical Grounding DINO}

The datasets used to fine-tune the medical grounding DINO is shown in Appendix Table \ref{tab:dino-combine_dataset}.


\begin{table*}[h!]
\centering
\small
\renewcommand{\arraystretch}{1.5} 
\begin{tabular}{m{1.3cm} m{1.3cm} m{1.3cm} m{1.3cm} m{8.5cm}}
\toprule
\textbf{Dataset} & \textbf{Modality} & \textbf{Anatomy} & \textbf{Image Number} & \textbf{Labels} \\ \midrule
WORD & CT & Abdomen & 9309 & Liver, Spleen, Kidney, Stomach, Gallbladder, Esophagus, Pancreas, Duodenum, Colon, Intestine, Adrenal, Rectum, Bladder, Head of femur \\ \hline
FLARE & CT & Abdomen & 4797 & Liver, Kidney, Spleen, Pancreas, Aorta, IVC, Adrenal Gland, Gallbladder, Esophagus, Stomach, Duodenum \\ \hline
VinDr-CXR & X-ray & Chest & 4394 & Aortic enlargement, Atelectasis, Calcification, Cardiomegaly, Consolidation, ILD, Infiltration, Lung Opacity, Nodule/Mass, Other lesion, Pleural effusion, Pleural thickening, Pneumothorax, Pulmonary fibrosis \\ \hline
MC & X-ray & Chest & 566 & Lung \\ \hline
BRATS & MRI & Brain & 14720 & Tumor \\ \hline
Cellseg & Histology & Cell & 229 & Cell \\ \bottomrule
\end{tabular}
\caption{Dataset overview for fine-tuning Grounding DINO.}
\label{tab:dino-combine_dataset}
\end{table*}

\section{Instruction Tuning Dataset Generation}

We represent our prompts for generating an instruction tuning dataset in Appendix Figure \ref{fig:appendix:gpt-instruction-tuning}.

\begin{figure*}[htp]
    \centering
    \includegraphics[width=\textwidth]{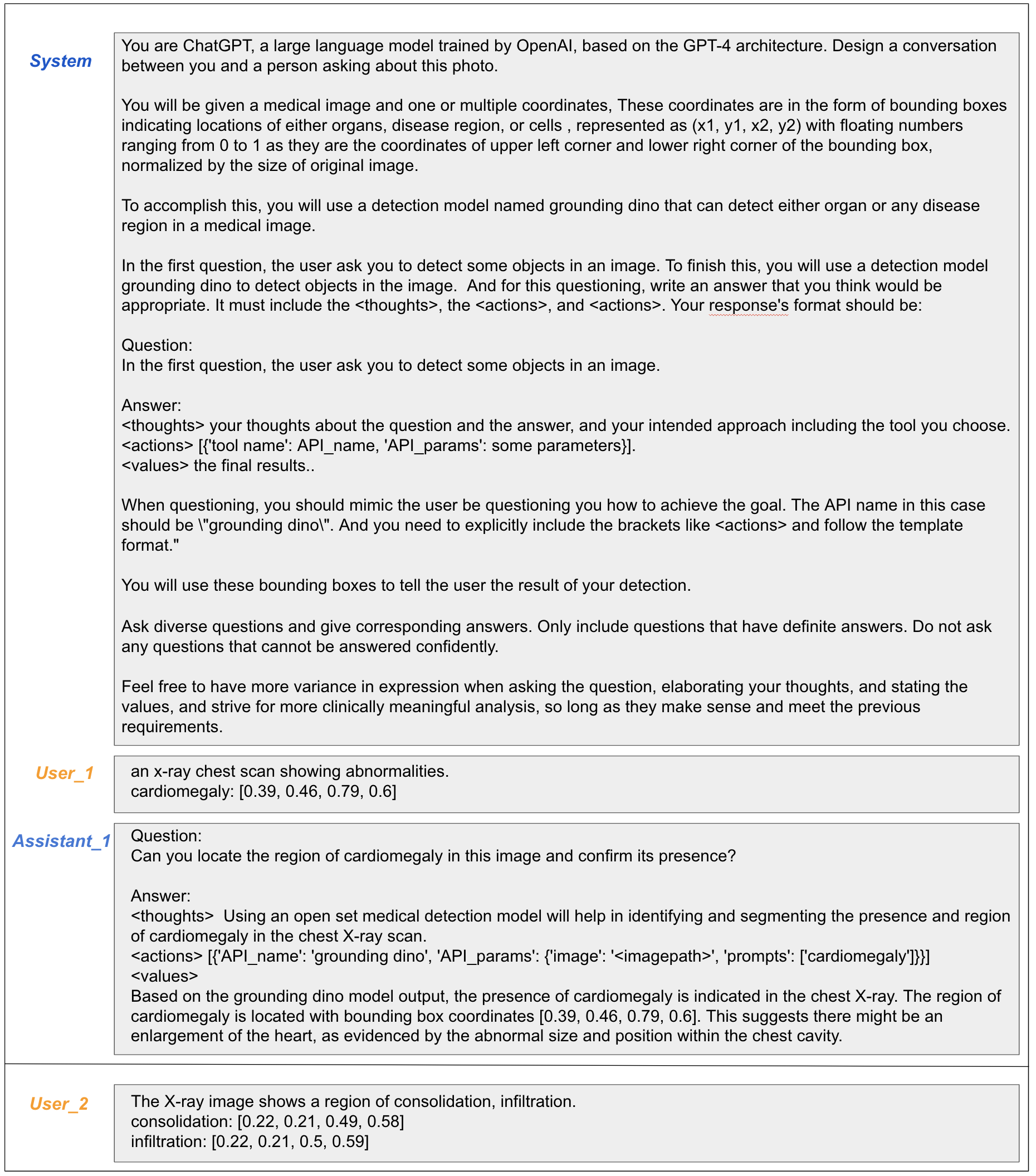}
    \caption{Pipeline of generating instruction-tuning dataset for the grounding task.}
    \label{fig:appendix:gpt-instruction-tuning}
\end{figure*}

\section{Agent Serving}
MMedAgent operates within the FastChat system
, which consists of web UI that interact with users, model workers hosting the language model, and various tools. A controller coordinates the activities between the web UI and model workers
. The entire system, including the 7B MMedAgent and all associated tools, can be run on an Nvidia A100 (80GB) GPU.
\begin{figure*}[htp]
    \centering
    \includegraphics[width=\textwidth]{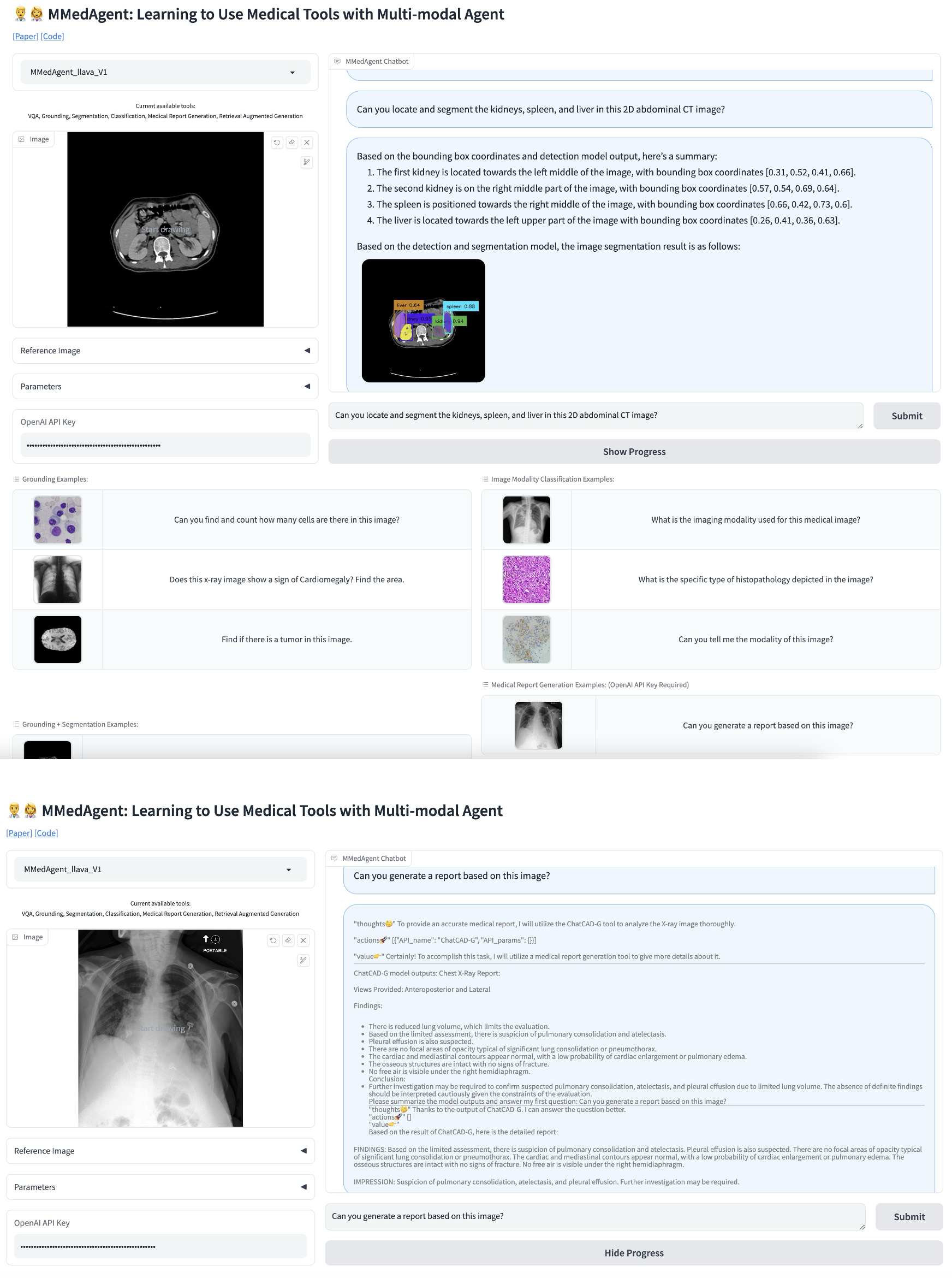}
    \caption{The user interface of the MMedAgent Web UI. Users can upload images and input questions in text, allowing MMedAgent to select the appropriate tool and provide comprehensive answers. The example shown demonstrates a request for segmenting organs in an abdominal CT image and generating diagnostic report for an X-ray image showing the thought progress of MMedAgent.}
    \label{fig:appendix:server}
    
\end{figure*}
\section{Evaluation Prompt}
We utilize GPT-4 to assess the answers generated by MMedAgent and other models with prompts shown in Appendix Figure \ref{fig:appendix:gpt-eval_prompt}.
\begin{figure*}[htp]
    \centering
    \includegraphics[width=\textwidth]{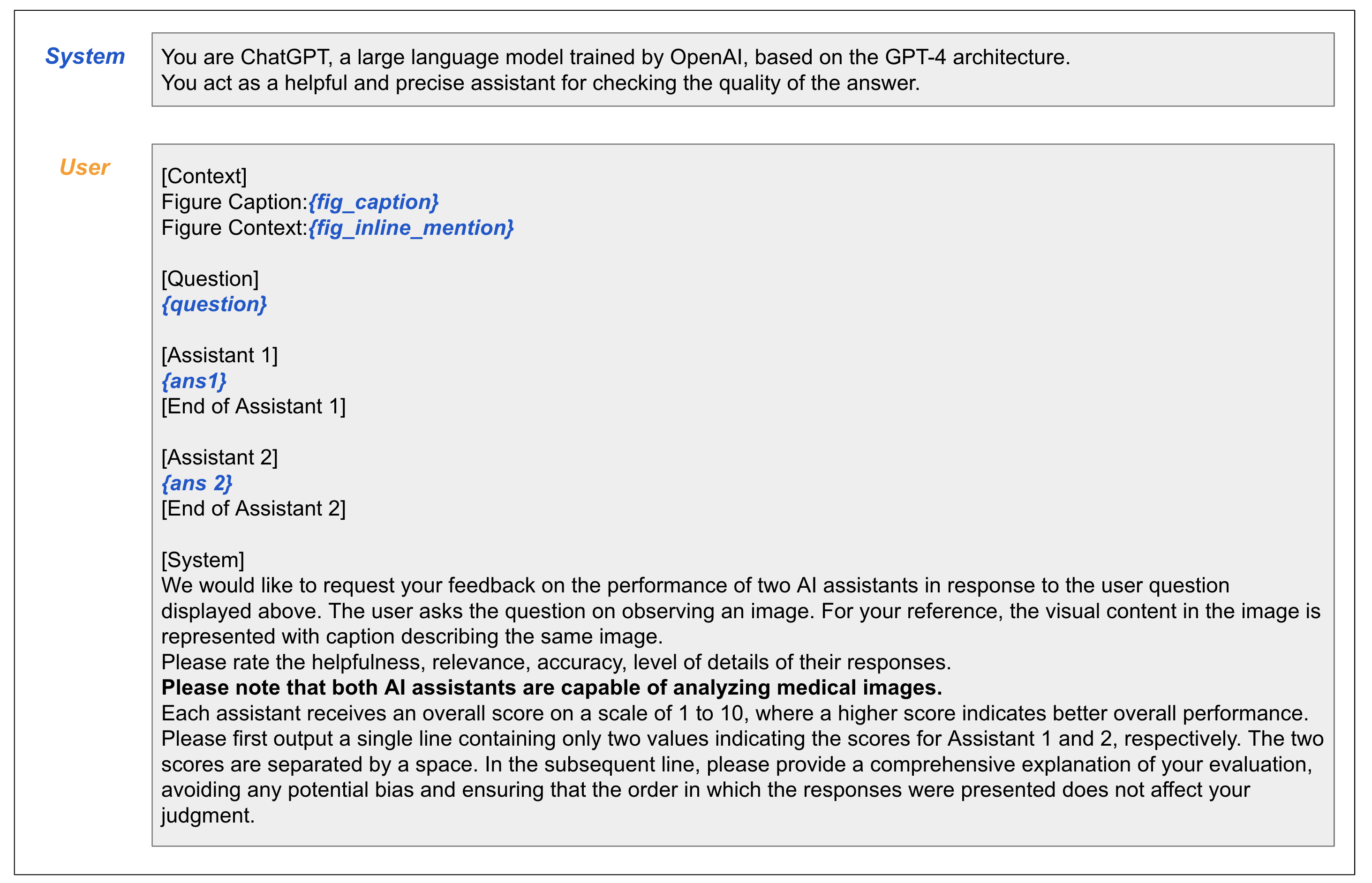}
    \caption{Evaluation pipeline. Assistant 1 is the model to be evaluated, which can be MMedAgent or LLaVA-Med and Assistant 2 is GPT-4o in our experiment.}
    \label{fig:appendix:gpt-eval_prompt}
    
\end{figure*}

\end{document}